\documentclass{article} 
\usepackage{iclr2027_conference,times}

\usepackage{amsmath,amsfonts,bm}

\def\eqref#1{equation~\ref{#1}}

\def\1{\bm{1}}

\DeclareMathAlphabet{\mathsfit}{\encodingdefault}{\sfdefault}{m}{sl}
\SetMathAlphabet{\mathsfit}{bold}{\encodingdefault}{\sfdefault}{bx}{n}

\usepackage[colorlinks, citecolor=teal, urlcolor=blue, linkcolor=black]{hyperref}

\usepackage{url}

\usepackage[utf8]{inputenc} 
\usepackage[T1]{fontenc}    
\usepackage{hyperref}       
\usepackage{url}            
\usepackage{booktabs}       
\usepackage{amsfonts}       
\usepackage{nicefrac}       
\usepackage{microtype}      
\usepackage{xcolor}         

\usepackage{algorithm}
\usepackage{algorithmic}
\usepackage{enumitem}
\usepackage{graphicx} 

\usepackage[table]{xcolor}
\usepackage{colortbl}
\usepackage{wrapfig}
\newcommand{\thickhline}{%
  \noalign {\hrule height 0.7pt} 
}
\usepackage{multirow}
\definecolor{myblue}{RGB}{100, 115, 160}

\usepackage{array}
\usepackage{bm}
\usepackage{xspace}
\usepackage{hyperref}
\usepackage[framemethod=TikZ]{mdframed}
\usepackage{pifont}
\usepackage{amsmath}
\usepackage{bbm}

\usepackage{caption}
\usepackage{array}
\usepackage{graphicx}
\usepackage{tabularx}
\usepackage{xcolor}

\usepackage[most]{tcolorbox}
\tcbuselibrary{listings,breakable}

\newtcblisting{promptbox}[1]{
  enhanced,
  breakable,
  listing only,
  listing options={
    basicstyle=\ttfamily\small,
    breaklines=true,
    columns=fullflexible,
    keepspaces=true
  },
  colback=gray!4,
  colframe=gray!55,
  boxrule=0.6pt,
  arc=2pt,
  left=6pt,
  right=6pt,
  top=8pt,
  bottom=8pt,
  title=\textbf{#1},
  fonttitle=\normalsize,
  coltitle=black,
  colbacktitle=gray!15,
  toptitle=2.5pt,
  bottomtitle=2.5pt
}

\definecolor{fuchsia}{rgb}{1.0, 0.0, 1.0}
\hypersetup{
  colorlinks=true,
  urlcolor=fuchsia
}

\newcommand{\descimg}[1]{%
\makebox[1.65cm][c]{%
\includegraphics[
  width=1.60cm,
  height=1.35cm,
  keepaspectratio
]{#1}%
}%
}

\newcommand{\qacell}[2]{%
\begin{minipage}[c]{3.65cm}
\centering
\setlength{\baselineskip}{8.8pt}
\underline{\textbf{Q:}}~#1\par
\vspace{1pt}
\underline{\textbf{A:}}~#2
\end{minipage}%
}

\newcommand{\optcell}[3]{%
\begin{minipage}[c]{3.65cm}
\centering
\setlength{\baselineskip}{8.8pt}
\underline{\textbf{Q:}}~#1\par
\vspace{1pt}
#2\par
\vspace{1pt}
\underline{\textbf{A:}}~#3
\end{minipage}%
}

\newcommand{\ourmethod}{{\fontfamily{lmtt}\selectfont \textbf{Switch-Reasoner}}\xspace}

\definecolor{myorange}{RGB}{233,144,61}

\title{Switch-Reasoner: Learn When to Think in Multitask Mixtures via Reinforcement Learning}

\author{Yiyang Fang$^{1,5}$ \; Pei Fu$^{2}$\footnotemark[1] \; Jinjie Li$^{3}$ \; Jian Liang$^{1}$ \; Wenke Huang$^{4}$ \; Ruijie Luo$^{1}$ \; \\ \textbf{Shaojie Zhang}$^{2}$ \; \textbf{Jian Luan}$^{2}$ \; \textbf{Yi R. (May) Fung}$^{5}$ \; \textbf{Mang Ye}$^{1}$\thanks{Corresponding author. Work done by Yiyang Fang during internship at Xiaomi Inc.} \\
$^{1}$ Wuhan University \;
$^{2}$ Xiaomi Inc \; 
$^{3}$ Wuhan University of Technology \; \\
$^{4}$ Nanyang Technological University \; \\
$^{5}$ The Hong Kong University of Science and Technology \\
\texttt{fangyiyang@whu.edu.cn, yemang@whu.edu.cn}
}

\iclrfinalcopy 



\begin{document}

\maketitle

\begin{abstract}
Multimodal Large Language Models (MLLMs) often follow a fixed  \emph{Think-then-Answer} paradigm, which is inefficient in heterogeneous multitask settings because simple inputs may not require explicit reasoning while difficult ones can benefit substantially from it. Learning when to think is also unstable during post-training, where imbalanced rollouts can drive the model toward always-thinking or always-direct behavior. We propose \ourmethod{}, a GRPO-based framework that learns to adaptively select reasoning modes for MLLMs. It treats thinking as a virtual tool invocation and allows the model to either answer directly or invoke explicit reasoning before answering. To stabilize this decision, we introduce a dual-level regulation mechanism that balances the overall use of \emph{Thinking Mode} and \emph{Direct Mode} while providing sample-level supervision based on the relative benefit of the two choices. Experiments on 11 multimodal tasks show that Switch-Reasoner reduces unnecessary reasoning while maintaining strong performance, achieving a better accuracy-efficiency trade-off. Code is available at \small\url{https://github.com/fuyyyyy/Switch-Reasoner}.
\end{abstract}

\section{Introduction}
\label{sec:intro}

Multimodal Large Language Models (MLLMs)~\citep{LLaVA15_CVPR23,li2024llava} have achieved remarkable progress in visual understanding and reasoning, supporting a broad range of tasks such as mathematical problem solving, scene understanding, chart reasoning, and visual question answering~\citep{chen2024internvl,li2024monkey,wang2024qwen2,SafetySurvey}. As these models become increasingly general-purpose, reinforcement learning (RL)-based post-training~\citep{liu2025reinforcement,zhou2025reinforced} has emerged as an effective approach for improving their reasoning capabilities beyond supervised fine-tuning~\citep{schulman2017proximal,rafailov2023direct}. In particular, Group Relative Policy Optimization (GRPO) has become widely adopted for reasoning-oriented training~\citep{DeepSeekR1_arXiv25,shao2024deepseekmath,GRPO_NeurIPS24} because of its simplicity and strong empirical performance without requiring a separately trained value model.

\begin{figure}[t]
  \centering
  \includegraphics[width=0.9\textwidth]{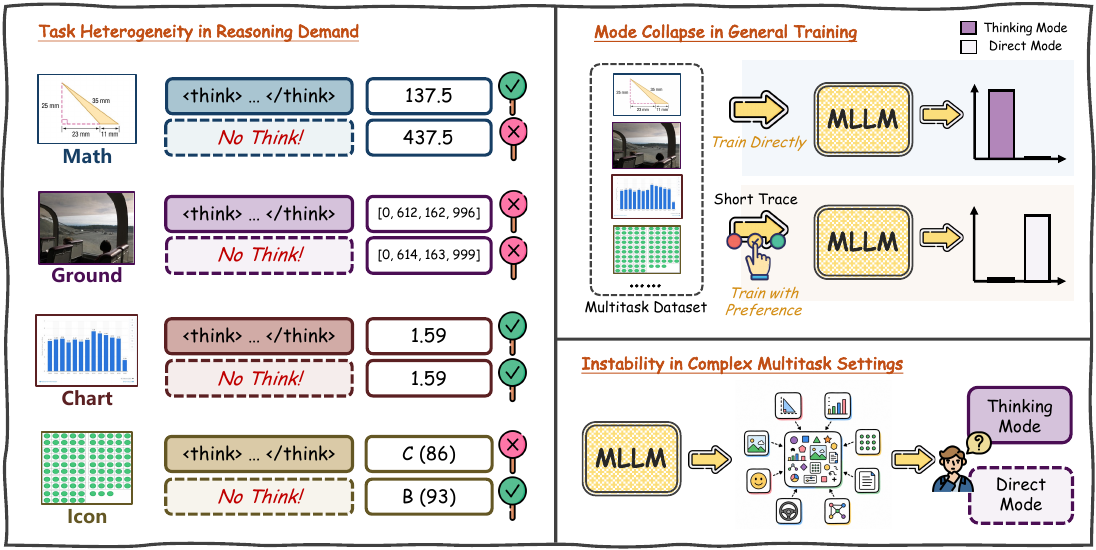}
  \vspace{-2pt}
  \caption{\textbf{Motivation for selective reasoning in heterogeneous multimodal tasks.}
  \textit{Left:} Tasks differ in their need for explicit reasoning.
  \textit{Top right:} Accuracy-only training can collapse to one mode.
  \textit{Bottom right:} Heterogeneous mixtures make stable  \emph{Thinking/Direct Mode} selection difficult.}
  \label{fig:motivation}
  \vspace{-10pt}
\end{figure}

However, improving reasoning accuracy alone is insufficient for practical MLLM deployment. Existing RL-based reasoning pipelines typically encourage a fixed \emph{Think-then-Answer} behavior~\citep{yao2025r1}, where the model generates an explicit chain of thought before producing its final answer. Although such deliberation is useful for difficult problems~\citep{feng2025video,huang2025boosting,rong2025safegrpo}, it incurs substantial token, latency, and deployment costs. More importantly, this strategy assumes that all inputs require the same amount of reasoning, which is fundamentally mismatched with heterogeneous multimodal tasks~\citep{wu2025more}. As illustrated in Figure~\ref{fig:motivation}, some tasks benefit substantially from explicit reasoning, whereas others can often be solved reliably through direct responses. A desirable MLLM should therefore selectively switch between a \emph{Thinking Mode} and a \emph{Direct Mode}, invoking reasoning only when necessary.

Recent work has pursued reasoning efficiency from two main directions. Early-exit~\citep{yang2025dynamic} and reasoning-pruning~\citep{hou2025thinkprune} methods reduce redundant computation by shortening chain-of-thought generation after reasoning has already begun~\citep{nagle2026terminator}. While effective at reducing output length, they do not address the preceding decision of whether explicit reasoning is needed at all. A more closely related line studies adaptive reasoning, allowing models to adaptively select reasoning behaviors across different inputs~\citep{fang2026thinkless, lou2025adacot, wang2026think, xu2026adapthink, huang2026learning}. To learn these behaviors, prior work employs constrained optimization and importance sampling~\citep{zhang2025adaptthink}, cold-start supervision~\citep{wang2026think, chen2025ares}, staged training and reward shaping~\citep{tu2026learning, zhu2025think, yang2025omni}, deterministic grouping~\citep{yang2025r}, and difficulty-aware curricula or rewards~\citep{xiao2026fast, chen2025ares}. Although effective in specific settings, these approaches often rely on carefully designed schedules or multi-stage pipelines, which can increase training complexity and cost. In addition, some studies primarily compare the relative merits of long versus short reasoning traces, providing useful empirical evidence but without explicitly learning a policy for deciding when reasoning is necessary~\citep{xiao2026fast, hou2025thinkprune}.

This challenge becomes particularly pronounced under GRPO training in heterogeneous multimodal mixtures. Transient reward advantages and imbalanced rollouts can drive the policy toward a dominant behavior~\citep{yue2026mars}: frequent rewards for deliberation encourage collapse into the \emph{Thinking Mode}, whereas overly strong penalties induce collapse into the \emph{Direct Mode}. Such mode collapse is especially problematic when the optimal reasoning strategy varies across tasks and individual instances~\citep{yang2026visionthink}. Therefore, the key problem we address is how to stably learn \textbf{instance-adaptive policy for selection between the  \emph{Thinking Mode} and the \emph{Direct Mode}} under GRPO training, while preventing collapse toward either excessive or insufficient deliberation.

To address this problem, we propose \ourmethod{}, a GRPO-based framework that learns to switch between the  \emph{Thinking Mode} and the \emph{Direct Mode} in heterogeneous multimodal mixtures. We first introduce a \textbf{Thinking-as-Tool formulation}, which represents explicit reasoning as a virtual tool invocation. Under a switchable prompting protocol, the model can either directly produce the final answer or invoke the thinking tool before answering, turning the whether-to-think decision into an explicit mode-selection action. To stabilize this mode-selection behavior during RL training, we further introduce a \textbf{dual-level mode-selection mechanism}. At the global level, it regulates the relative usage of the two modes to avoid collapse toward a dominant behavior; at the sample level, it provides mode-selection supervision based on the relative benefit of thinking and direct answering for each input. Together, these components encourage balanced exploration during training and instance-adaptive allocation of reasoning computation.

The main contributions of this work are summarized as follows:
\begin{itemize}
    \item We introduce a Thinking-as-Tool formulation with a switchable prompting protocol, turning the whether-to-think decision into an explicit and learnable routing action before generating long reasoning traces.

    \item We develop a dual-level mode regulation mechanism that combines global balance control with sample-level counterfactual route supervision, preventing mode collapse while encouraging instance-adaptive reasoning allocation.

    \item Extensive experiments on heterogeneous multimodal benchmarks show that \ourmethod{} achieves a better accuracy--efficiency trade-off
    .
\end{itemize}

\section{Related Work}

\subsection{Reinforcement Learning for MLLMs}
Reinforcement learning has become an effective approach for improving the reasoning capabilities of multimodal large language models (MLLMs)~\citep{liu2025reinforcement,zhou2025reinforced}. Among existing methods, Group Relative Policy Optimization (GRPO)~\citep{shao2024deepseekmath,GRPO_NeurIPS24} is widely adopted because it estimates relative advantages from multiple sampled outputs without requiring a separate value model~\citep{fang2026emo}, offering a simple and effective alternative to PPO~\citep{schulman2017proximal}. DeepSeek-R1~\citep{DeepSeekR1_arXiv25} further demonstrated the effectiveness of GRPO for reasoning-oriented post-training in large language models. Recent work has extended GRPO to multimodal settings, including visual reasoning, video understanding, and omni-modal perception. Text-Debiased Hint-GRPO~\citep{huang2025boosting} reduces linguistic bias in multimodal reasoning, R1-VL~\citep{zhang2025r1} introduces step-wise optimization for vision-language learning, and R1-Omni~\citep{zhao2025r1} applies reinforcement learning to omni-modal emotion recognition. Video-R1~\citep{feng2025video}, VideoChat-R1~\citep{li2025videochat}, and Visual-RFT~\citep{liu2025visual} further demonstrate the scalability of RL-based post-training across video and vision-centric tasks. Different from these works, which primarily improve task performance through general-purpose RL objectives, our work focuses on learning when explicit reasoning is necessary in heterogeneous multimodal mixtures and stabilizing this behavior under GRPO training.

\subsection{Efficient Reasoning in MLLMs}

Recent work on efficient reasoning in LLMs reduces unnecessary computation either by shortening reasoning traces after generation or by adaptively determining whether and how deeply to deliberate~\citep{huang2025adactrl}. Early-exit methods terminate reasoning trajectories once sufficient evidence is available~\citep{yang2025dynamic, nagle2026terminator}, while pruning methods compress redundant chain-of-thought content~\citep{hou2025thinkprune}. Other studies enable adaptive reasoning through explicit think-or-direct decisions~\citep{fang2026thinkless, zhang2025adaptthink, lou2025adacot}, staged optimization~\citep{tu2026learning, zhu2025think}, or step-wise budget allocation~\citep{wang2026adaptr1}. Building on these advances, recent work extends adaptive reasoning to MLLMs, where visual, audio, and heterogeneous multimodal inputs introduce additional variation in reasoning demand. Selective reasoning has been explored in vision-language models~\citep{wang2026think, yang2025r}, difficulty-aware multimodal training~\citep{xiao2026fast, chen2025ares}, and omni-modal reasoning settings~\citep{yang2025omni, wu2026audio}. Other methods consider multi-path visual reasoning~\citep{huang2026learning, li2025mixture}, adaptive inference computation~\citep{xu2025learning}, and efficient reasoning for video or tool-augmented visual tasks~\citep{liu2026videoauto, wang2026adatooler}. In contrast, our work focuses on stably learning instance-adaptive selection between \emph{Thinking Mode} and \emph{Direct Mode} in heterogeneous multimodal mixtures under GRPO.

\section{Switch-Reasoner}

\subsection{Preliminaries}

Group Relative Policy Optimization (GRPO) is a variant of Proximal Policy Optimization (PPO) that can improve mathematical, visual, and other multimodal reasoning abilities. It maintains a current policy $\pi_{\theta}$ and an old policy $\pi_{\text{old}}$ from the previous iteration. Given a prompt $q \sim \rho_Q$, GRPO samples a group of outputs $o_1,o_2,\ldots,o_G$ from $\pi_{\text{old}}$ and optimizes $\pi_{\theta}$ by maximizing:
\begin{equation}
\label{eq:grpo}
\mathcal{J}_{\mathrm{GRPO}}(\theta) 
= \mathbb{E}_{q \sim \rho_Q} \mathbb{E}_{o \sim \pi_{\text{old}}(\cdot|q)} 
\Bigg[
\frac{1}{G} \sum_{i=1}^{G} 
f_{\epsilon}\!\left(
\frac{\pi_{\theta}(o_i|q)}{\pi_{\text{old}}(o_i|q)}, \hat{A}_i
\right)
\Bigg] - \beta\, \mathbb{D}_{KL}\!\left[\pi_{\theta} \| \pi_{\text{ref}}\right],
\end{equation}
where $\beta$ is the KL regularization hyperparameter, and $f_\epsilon(x,y)=\min(xy,\text{clip}(x,1-\epsilon,1+\epsilon)y)$. $\hat{A}_i$ denotes the advantage of the $i$-th output, computed from relative rewards within the sampled group.

For each question $q$, a group of outputs $\{o_1,o_2,\ldots,o_G\}$ is sampled from $\pi_{\text{old}}$. A reward function $\mathcal{R}$ scores these outputs, yielding $\mathbf{r}=\{r_1,r_2,\ldots,r_G\}$, where $r_i=\mathcal{R}(q,o_i)$. The mean reward is $\mu=\frac{1}{G}\sum_{i=1}^{G}r_i$, and the standard deviation is $\sigma=\sqrt{\frac{1}{G}\sum_{i=1}^{G}(r_i-\mu)^2}$. The normalized advantage of the $i$-th rollout is $\hat{A}_i=\frac{r_i-\mu}{\sigma}$. This normalization gives advantages zero mean and unit variance within each group, stabilizing gradients and promoting consistent optimization.

\begin{figure}[t]
  \centering
  \includegraphics[width=1.0\textwidth]{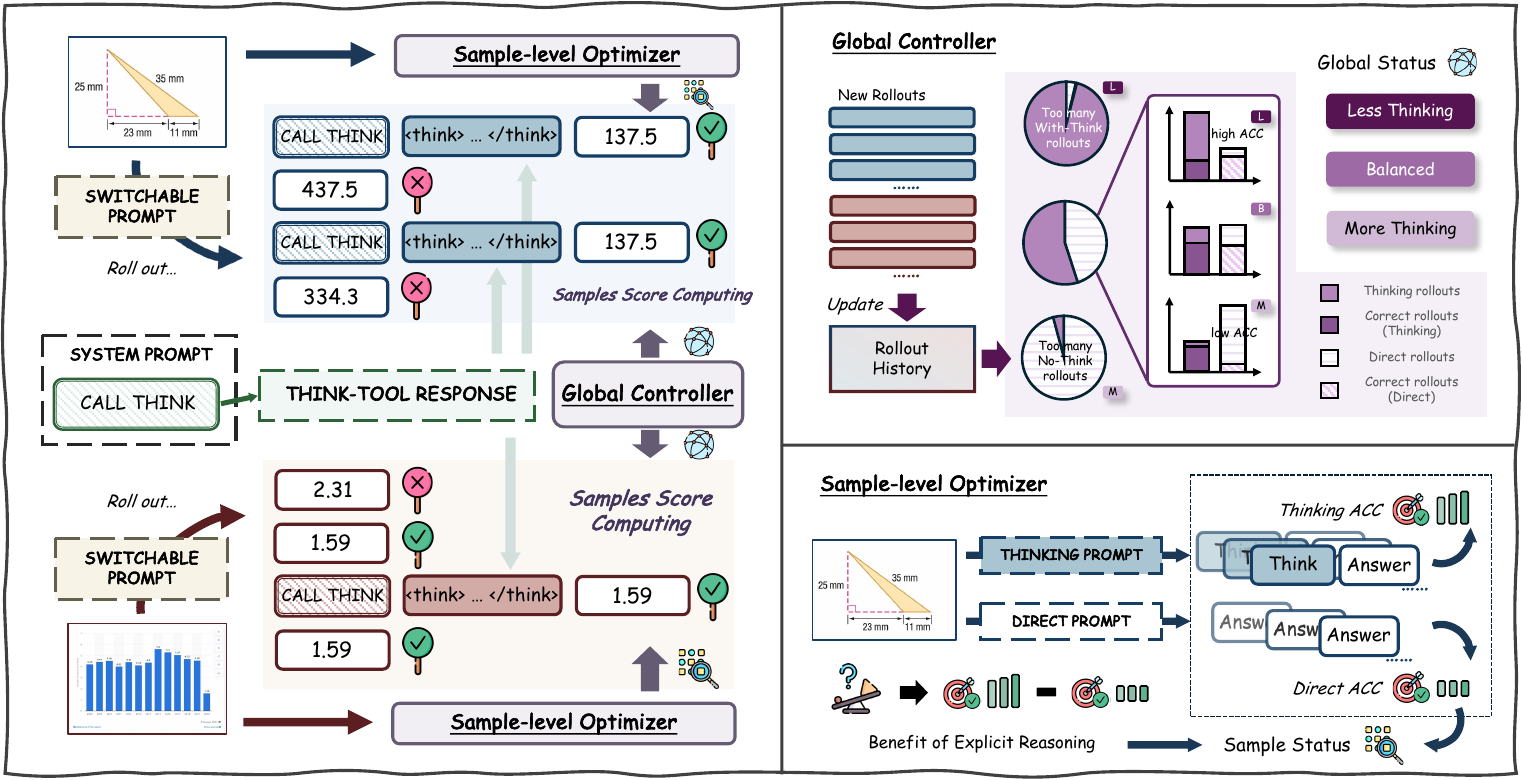}
  \vspace{-10pt}
  \caption{\textbf{Overall framework of \ourmethod{}.} The model first selects between \emph{Direct Mode} and \emph{Thinking Mode} through a switchable prompt. Tool invocation triggers explicit reasoning, while direct answers bypass it. During training, a global controller regulates mode balance, and a sample-level optimizer provides counterfactual route supervision.}
  \label{fig:framework}
  \vspace{-10pt}
\end{figure}

\subsection{Thinking-as-Tool Switchable Prompt}

To enable selective reasoning in MLLMs, we reformulate  \emph{Thinking Mode} as an explicit tool invocation rather than an implicit always-on generation pattern. The key idea is to expose whether explicit reasoning is performed as an observable and optimizable action, allowing the model to determine when additional reasoning is necessary.

We introduce a virtual reasoning tool $T_{\mathrm{think}}$ with a switchable prompting protocol. Given a multimodal input $q=(x_v,x_t)$, the model first generates an initial response under a constrained format with two possible behaviors: directly producing the final answer, referred to as \emph{Direct Mode}, or emitting a tool-call token to invoke $T_{\mathrm{think}}$, referred to as \emph{Thinking mode}. This design turns the reasoning-mode selection decision into a discrete and observable action.

When the thinking tool is invoked, a follow-up think-tool response switches the model into  \emph{Thinking Mode} and requires intermediate reasoning before the final answer. Otherwise, the model outputs the answer directly. During rollout, the model first follows the switchable protocol; a tool call continues the trajectory with the follow-up think-tool response, while a direct answer is treated as a Direct-answering-mode trajectory. Generation can therefore be viewed as an implicit two-stage process: mode selection followed by either direct answering or explicit reasoning.

Formally, let $m\in{0,1}$ denote the mode decision, where $m=1$ invokes the thinking tool and $m=0$ denotes \emph{Direct Mode}. The policy is factorized as
\begin{equation}
\label{eq:policy_factorization}
\pi_\theta(o,m|q) = \pi_\theta(m|q)\pi_\theta(o|q,m),
\end{equation}
where $o$ denotes the final output sequence. Although written in this form, the mode decision is realized implicitly through the output format and is influenced by the input question, visual information, and decoding context.

Compared with conventional  \emph{Think-then-Answer} prompting, this formulation allows simple samples to bypass unnecessary reasoning while providing a discrete routing interface that can be optimized directly under reinforcement learning. All prompt details are provided in Appendix~\ref{ap:prompt}.

\subsection{Global Mode Balance Control}

Although Thinking-as-Tool enables explicit switching between  \emph{Thinking Mode} and \emph{Direct Mode}, a switchable interface alone does not guarantee stable mode preference during reinforcement learning. Short-term reward advantages and sampling fluctuations may cause the policy to overuse one mode. Explicit reasoning can temporarily obtain higher rewards through longer trajectories, whereas \emph{Direct Mode} may dominate when it is more efficient or stable at certain stages.

To address this issue, we introduce \textbf{Global Mode Balance Control}, an online delayed controller that provides a batch-level signal for regulating the relative use of the two modes. At each training iteration, the controller applies the signal estimated from the previous rollout batch to the current reward computation, and then updates its state using the statistics of the current batch. Specifically, after reward evaluation, we collect format-valid trajectories and compute the number and accuracy of \emph{Thinking Mode} and \emph{Direct Mode} rollouts: $n_{\mathrm{think}}$, $n_{\mathrm{direct}}$, $a_{\mathrm{think}}$, and $a_{\mathrm{direct}}$. Their usage ratios are
\begin{equation}
\label{eq:usage_ratio}
p_{\mathrm{direct}} = \frac{n_{\mathrm{direct}}}{n_{\mathrm{think}}+n_{\mathrm{direct}}},
\qquad
p_{\mathrm{think}} = \frac{n_{\mathrm{think}}}{n_{\mathrm{think}}+n_{\mathrm{direct}}}.
\end{equation}
Only format-valid trajectories are used because malformed outputs cannot reliably reflect the actual reasoning-mode decision.

Let
\begin{equation}
\label{eq:accuracy_gap}
\Delta a = a_{\mathrm{direct}} - a_{\mathrm{think}}
\end{equation}
denote the accuracy advantage of \emph{Direct Mode}. We generate a global control signal $B$ according to performance difference and usage ratio:
\begin{equation}
\label{eq:global_control}
B =
\begin{cases}
+B_g, & \Delta a>\delta \ \text{and}\ p_{\mathrm{direct}}<\tau_{\mathrm{low}},\\
-B_g, & -\Delta a>\delta \ \text{and}\ p_{\mathrm{direct}}>\tau_{\mathrm{high}},\\
0, & \text{otherwise}.
\end{cases}
\end{equation}

When $B>0$, \emph{Direct Mode} performs better but is underused, so the training reward relatively favors direct answers by penalizing Thinking-mode trajectories. When $B<0$,  \emph{Thinking Mode} performs better while \emph{Direct Mode} is overused, so the training reward relatively favors reasoning by penalizing Direct-answering-mode trajectories.

To preserve exploration when one mode becomes excessively dominant, we introduce an extreme imbalance safeguard:
\begin{equation}
\label{eq:imbalance_control}
B =
\begin{cases}
+B_f, & p_{\mathrm{think}}>\pi_{\mathrm{high}},\\
-B_f, & p_{\mathrm{think}}<\pi_{\mathrm{low}}.
\end{cases}
\end{equation}
A count-ratio criterion can additionally identify nearly single-mode batches and restore exploration when one route almost disappears. The updated control signal is stored and used in the next training iteration. This one-batch delayed design avoids circular dependence between the rewards of the current batch and the statistics computed from those same rewards, while still providing a coarse-grained routing preference based on empirical performance and mode usage.

\subsection{Sample-level Fine-grained Optimization}

Global mode control provides only a batch-level preference, whereas the need for explicit reasoning is inherently sample-dependent. Even within the same batch, some inputs can be solved reliably with concise direct responses while others require multi-step visual or textual reasoning. We therefore introduce \textbf{Sample-level Fine-grained Optimization} to provide counterfactual route supervision for each training sample.

For each input $q_i$, we evaluate the current policy under two fixed choices: \emph{Direct Mode} and  \emph{Thinking Mode}. Specifically, we construct a direct prompt that forces an answer without invoking the thinking tool and a thinking prompt that requires explicit reasoning before the final answer:
\begin{equation}
\label{eq:counterfactual_rollout}
\{o_{i,k}^{\mathrm{direct}}\}_{k=1}^{K} \sim \pi_\theta(\cdot|q_i,m=0),
\qquad
\{o_{i,k}^{\mathrm{think}}\}_{k=1}^{K} \sim \pi_\theta(\cdot|q_i,m=1).
\end{equation}

Let $R_{\mathrm{acc}}(q,o)$ denote the accuracy component of the task reward. We estimate the empirical success rate of each route as
\begin{equation}
\label{eq:direct_success}
s_i^{\mathrm{direct}} = \frac{1}{K}\sum_{k=1}^{K}\mathbbm{1}\left[R_{\mathrm{acc}}(q_i,o_{i,k}^{\mathrm{direct}})>0\right],
\end{equation}
\begin{equation}
\label{eq:think_success}
s_i^{\mathrm{think}} = \frac{1}{K}\sum_{k=1}^{K}\mathbbm{1}\left[R_{\mathrm{acc}}(q_i,o_{i,k}^{\mathrm{think}})>0\right].
\end{equation}

Their difference reflects the sample-specific benefit of explicit reasoning:
\begin{equation}
\label{eq:sample_gain}
\Delta_i = s_i^{\mathrm{think}} - s_i^{\mathrm{direct}}.
\end{equation}

Based on this comparison, each sample is assigned to one of three categories:
\begin{equation}
\label{eq:sample_category}
c_i =
\begin{cases}
\texttt{must-think}, & \Delta_i\geq\delta_{\mathrm{think}},\\
\texttt{safe-direct}, & s_i^{\mathrm{direct}}\geq\tau_{\mathrm{direct}} \ \text{and}\ \Delta_i\leq -\delta_{\mathrm{safe}},\\
\texttt{uncertain}, & \text{otherwise}.
\end{cases}
\end{equation}

\texttt{must-think} indicates that  \emph{Thinking Mode} provides a clear improvement. In contrast, \texttt{safe-direct} is assigned only when \emph{Direct Mode} is already reliable and  \emph{Thinking Mode} does not provide a sufficient benefit; in our default setting, this requires the empirical thinking gain to fall below a negative safety margin. For uncertain samples, we impose no strong preference to avoid introducing noisy route supervision.

Together, the global controller captures the overall tendency of the two modes, while sample-level comparison identifies the inputs that benefit from reasoning and those that can be answered directly.

\subsection{Reward Computation}

We design the reward to jointly encourage answer correctness, format compliance, balanced mode usage, and sample-level route selection. For each response $o_i$, the base reward is
\begin{equation}
\label{eq:base_reward}
r_i^{\mathrm{base}}=(1-\alpha)r_i^{\mathrm{acc}}+\alpha r_i^{\mathrm{fmt}},
\end{equation}
where $r_i^{\mathrm{acc}}$ measures task correctness and $r_i^{\mathrm{fmt}}$ measures compliance with the switchable protocol.

We apply the global control signal from the previous rollout batch as a multiplicative penalty on the temporarily disfavored mode:
\begin{equation}
\label{eq:global_reward}
r_i^{\mathrm{global}}
=
r_i^{\mathrm{base}}
\cdot
\eta^{\mathbbm{1}\left[(B>\epsilon \land m_i=1)\lor(B<-\epsilon \land m_i=0)\right]},
\end{equation}
where $m_i=1$ denotes  \emph{Thinking Mode}, $m_i=0$ denotes \emph{Direct Mode}, and $\eta\in(0,1)$ is the penalty factor. The controller state $B$ is updated after scoring the current batch and used in the next iteration.

For sample-level route supervision, we define a preferred-route sign
\begin{equation}
\gamma_i =
\begin{cases}
+1, & c_i=\texttt{must-think},\\
-1, & c_i=\texttt{safe-direct},\\
0, & c_i=\texttt{uncertain},
\end{cases}
\end{equation}
and a selected-route sign
\begin{equation}
\mu_i =
\begin{cases}
+1, & m_i=1,\\
-1, & m_i=0.
\end{cases}
\end{equation}
The route reward is then
\begin{equation}
\label{eq:route_reward}
r_i^{\mathrm{route}} = w_{c_i}\gamma_i\mu_i,
\end{equation}
where $w_{c_i}=w_{\mathrm{think}}$ for \texttt{must-think}, $w_{c_i}=w_{\mathrm{direct}}$ for \texttt{safe-direct}, and $w_{c_i}=0$ for \texttt{uncertain}. Thus, the route reward is positive when the selected route matches the preferred route, negative when it mismatches, and zero for uncertain samples.

The final reward is
\begin{equation}
\label{eq:final_reward}
r_i^\ast=r_i^{\mathrm{global}}+r_i^{\mathrm{route}}.
\end{equation}
This reward is used for group-wise advantage estimation in GRPO.

The relevant notation table and algorithms are provided in Appendix~\ref{ap:notation} and Appendix~\ref{ap:algorithm}.

\definecolor{mygray}{gray}{.9}
\definecolor{myyellow}{RGB}{252, 240, 199}
\definecolor{myblue}{RGB}{100, 115, 160}

\begin{table}[t]
	\centering
 	\caption{\textbf{Comparison on multitask multimodal benchmarks.} Reported results are averaged over the last three validation steps, except for Vanilla. \textcolor{myblue}{$\pm$} denotes the standard deviation. * indicates the variant without sample-level fine-grained correction. See Section~\ref{sec:compare} for details.}
    \label{tab:sota}
    \vspace{-0.2cm}
    \setlength{\tabcolsep}{5.5pt}
    \renewcommand\arraystretch{1.2}
	\scalebox{0.92}{
	\begin{tabular}{c||c|c|c|c|cc}
		\hline\thickhline
        \rowcolor{mygray}
        \rowcolor{mygray}
        & & & & & \multicolumn{2}{c}{Overall} \\
        \rowcolor{mygray}
        \multirow{-2}{*}{Method} & \multirow{-2}{*}{Mathematics} & \multirow{-2}{*}{Chart/Doc} & \multirow{-2}{*}{Grounding} & \multirow{-2}{*}{General} & Score & Think Rate \\
		\hline
		\hline
        \multicolumn{7}{l}{\textcolor{gray!80}{\textit{Qwen3-VL-4B}}}\\
        \rowcolor{gray!8}
        Vanilla & 26.20 & 91.87 & 57.60 & 40.43 & 48.76 & 0.00\% \\
        
        GRPO-Thinking & 71.90 & 91.52 & 61.40 & 64.84 & 70.66 \textcolor{myblue}{$\pm$ 0.54} & 100.00\% \\
        \rowcolor{gray!8}
        GRPO-Direct & 43.40 & 91.47 & 61.07 & 64.08 & 65.02 \textcolor{myblue}{$\pm$ 0.23} & 0.00\% \\
        
        \ourmethod{}* & 66.83 & 92.06 & 62.13 & 66.47 & 70.79 \textcolor{myblue}{$\pm$ 0.68} & 85.73\% \\
        \rowcolor{gray!8}
        \ourmethod{} & 67.23 & 91.66 & 60.27 & 68.26 & \textbf{71.60} \textcolor{myblue}{$\pm$ 0.30} & 51.53\% \\
        \hline
        \multicolumn{7}{l}{\textcolor{gray!80}{\textit{Qwen3-VL-8B}}}\\
        \rowcolor{gray!8}
        Vanilla & 29.90 & 92.22 & 58.60 & 42.16 & 50.53 & 0.00\% \\
        GRPO-Thinking & 73.57 & 93.01 & 60.13 & 66.19 & 71.86 \textcolor{myblue}{$\pm$ 0.34} & 100.00\% \\
        \rowcolor{gray!8}
        GRPO-Direct & 48.23 & 92.30 & 58.87 & 64.03 & 65.83 \textcolor{myblue}{$\pm$ 0.07} & 0.00\% \\
        
        \ourmethod{}* & 68.43 & 92.67 & 58.20 & 68.60 & 72.00 \textcolor{myblue}{$\pm$ 0.24} & 30.25\% \\
        \rowcolor{gray!8}
        \ourmethod{} & 68.80 & 92.62 & 58.93 & 68.78 & \textbf{72.22} \textcolor{myblue}{$\pm$ 0.16} & 37.73\% \\
		\hline
	\end{tabular}}
    \vspace{-6pt}
\end{table}

\begin{figure}[t]
  \centering
  \includegraphics[width=1.0\textwidth]{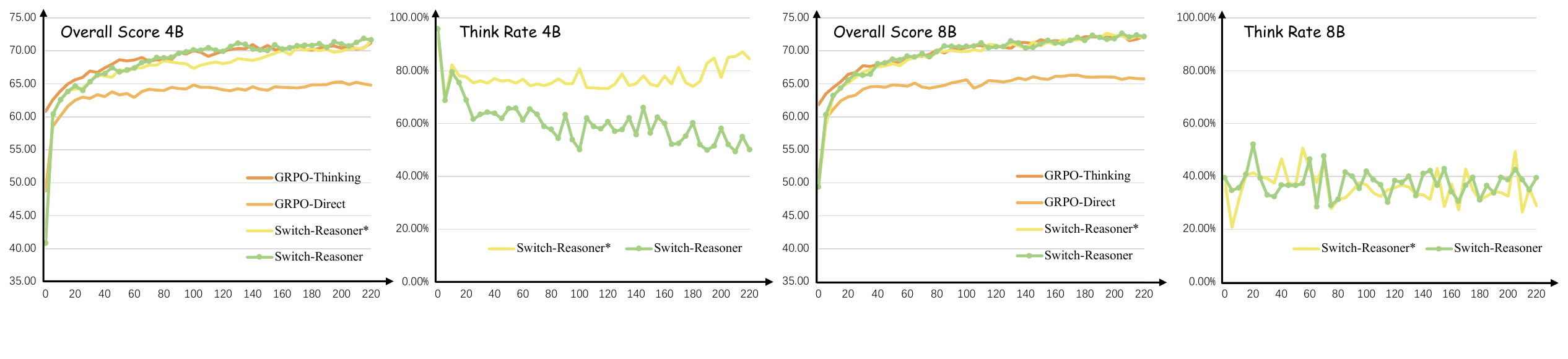}
  \vspace{-31pt}
  \caption{Training dynamics of validation performance and thinking rate. Refer to Section~\ref{sec:visual}.}
  \label{fig:val_curves}
  \vspace{-11pt}
\end{figure}

\section{Experiment}
\subsection{Experimental Setup}
\label{sec:exp_setup}

\noindent\textbf{Dataset and Evaluation Metrics.}
We conduct experiments on a heterogeneous multitask benchmark with 11 representative multimodal tasks, covering mathematical and visual reasoning, chart/document understanding, localization, and general multimodal understanding. We follow the standard metric of each task, and more details are provided in the Appendix~\ref{ap:dataset}.

\noindent\textbf{Architecture and Counterparts.}
We use Qwen3-VL-4B-Instruct and Qwen3-VL-8B-Instruct as vanilla base models~\citep{bai2025qwen3} for RL training. We compare our method with Vanilla model, all-thinking and all-direct variants, as well as a variant without sample-level fine-grained correction, to evaluate the effects of switchable reasoning and instance-level route supervision.

\noindent\textbf{Implementation Details.}
All models are trained with GRPO using EasyR1.\footnote{%
\begingroup
\hypersetup{urlcolor=black}
\url{https://github.com/hiyouga/easyr1}%
\endgroup} In the switchable reasoning setting, each prompt allows the model either to directly produce the final answer or to invoke the thinking tool before responding. We set the maximum prompt length to 8,192 tokens and train all models for 20 epochs using 8 NVIDIA H20 GPUs, each equipped with 96 GB of memory. Additional hyperparameter details and parameter ablation results are provided in Appendix~\ref{ap:hyper}.

\begin{figure}[t]
  \centering
  \includegraphics[width=1.0\textwidth]{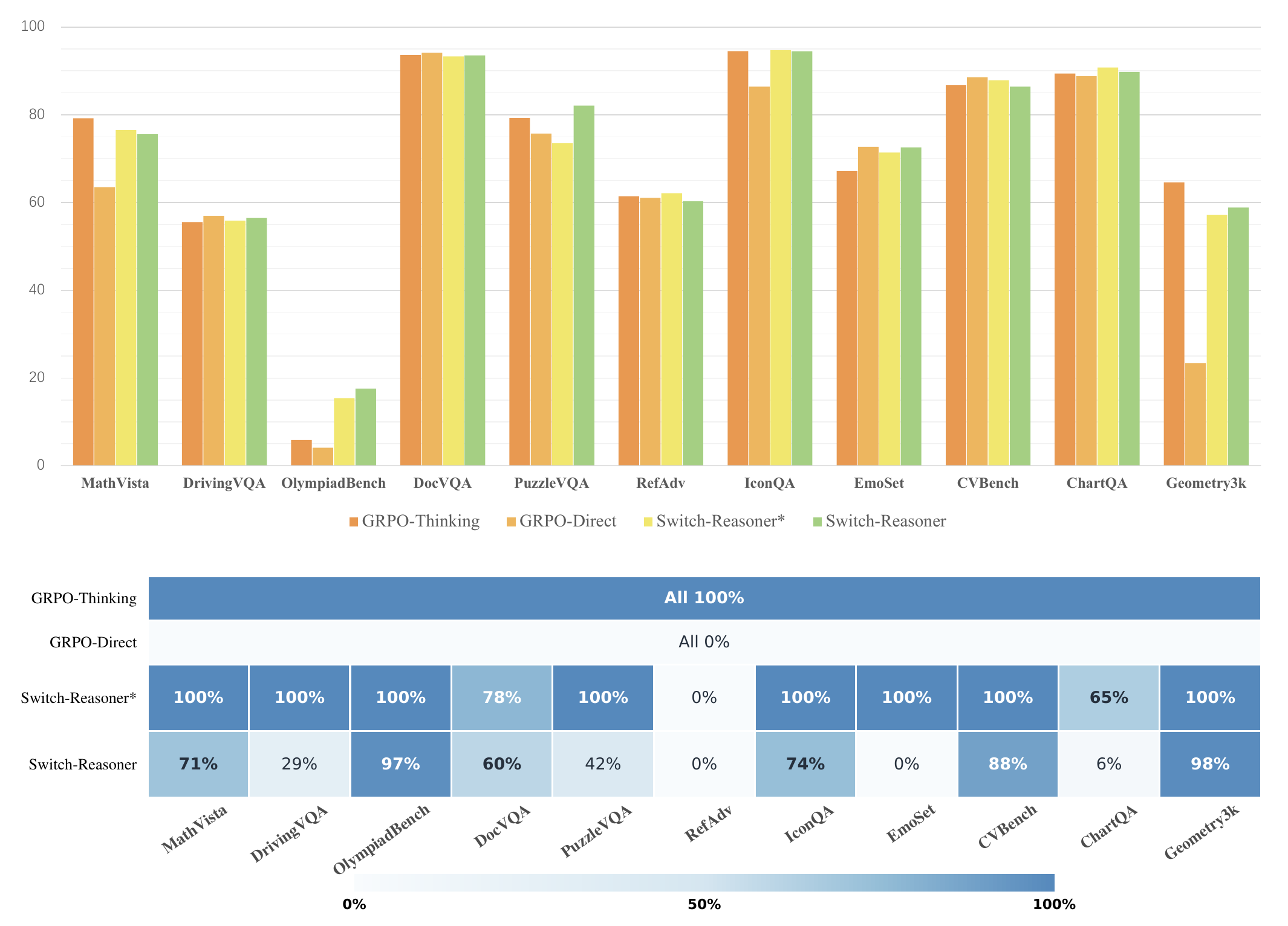}
  \vspace{-17pt}
  \caption{Dataset-level score and thinking rate of the 4B model. See Section~\ref{sec:visual} for details.}
  \label{fig:dataset_acc}
  \vspace{-10pt}
\end{figure}

\subsection{Comparison Experiment}
\label{sec:compare}
Table~\ref{tab:sota} compares different training strategies across four categories of multimodal tasks. The reported results are averaged over the last three validation steps. GRPO-Thinking achieves strong performance but requires a 100\% thinking rate, while GRPO-Direct eliminates explicit thinking entirely, leading to clear performance degradation.

Our method achieves a better trade-off between performance and thinking cost, with particularly clear gains on Qwen3-VL-4B. Switch-Reasoner reaches the best overall score among all 4B methods, improving over both GRPO-Thinking and GRPO-Direct while reducing the thinking rate to 51.53\%. On Qwen3-VL-8B, Switch-Reasoner also achieves the best overall score, reaching 72.22 with only a 37.73\% thinking rate. These results indicate that the proposed switching strategy can effectively reduce unnecessary reasoning while preserving, and in some cases improving, multimodal performance. The improvement is especially evident on the General task, which contains a more diverse mixture of datasets and therefore better reflects the benefit of adaptive reasoning-mode selection.

The comparison between Switch-Reasoner* and Switch-Reasoner further demonstrates the benefit of sample-level fine-grained correction. Compared with the variant without this correction, the full method achieves a higher overall score with a comparable or lower thinking rate. This suggests that fine-grained route supervision helps the model make more reliable sample-level switching decisions.

\subsection{Visualization Analysis}
\label{sec:visual}

We conduct visualization experiments to further analyze the behavior of Switch-Reasoner during training and evaluation. Specifically, we visualize the validation performance curves, dataset-level score, and dataset-level thinking rates to understand how the model learns to allocate reasoning computation across different tasks.

\noindent\textbf{Visualization of Training Dynamics.}
To analyze the training behavior of different reasoning strategies, we visualize the validation score and thinking rate across training steps. As shown in Figure~\ref{fig:val_curves}, Switch-Reasoner gradually learns to invoke thinking selectively. It achieves higher validation scores than GRPO-Direct while using substantially fewer thinking trajectories than GRPO-Thinking. This indicates that our method does not simply suppress reasoning uniformly, but learns an adaptive reasoning policy that balances task performance and thinking cost.

\noindent\textbf{Visualization of Dataset-level Performance.}
We further visualize dataset-level score and thinking rates to examine how different methods behave across heterogeneous tasks. As shown in Figure~\ref{fig:dataset_acc}, GRPO-Thinking consistently relies on explicit reasoning for all datasets, whereas GRPO-Direct always produces direct answers without thinking. In contrast, Switch-Reasoner assigns different thinking rates to different datasets while maintaining competitive or stronger accuracy. Reasoning-intensive datasets tend to receive higher thinking rates, while some recognition- or understanding-oriented datasets can be handled effectively with direct answering. These results show that Switch-Reasoner learns to adapt its reasoning mode according to dataset-level demands, reducing unnecessary thinking while preserving overall performance.

\begin{figure}
  \centering
  \includegraphics[width=1.0\textwidth]{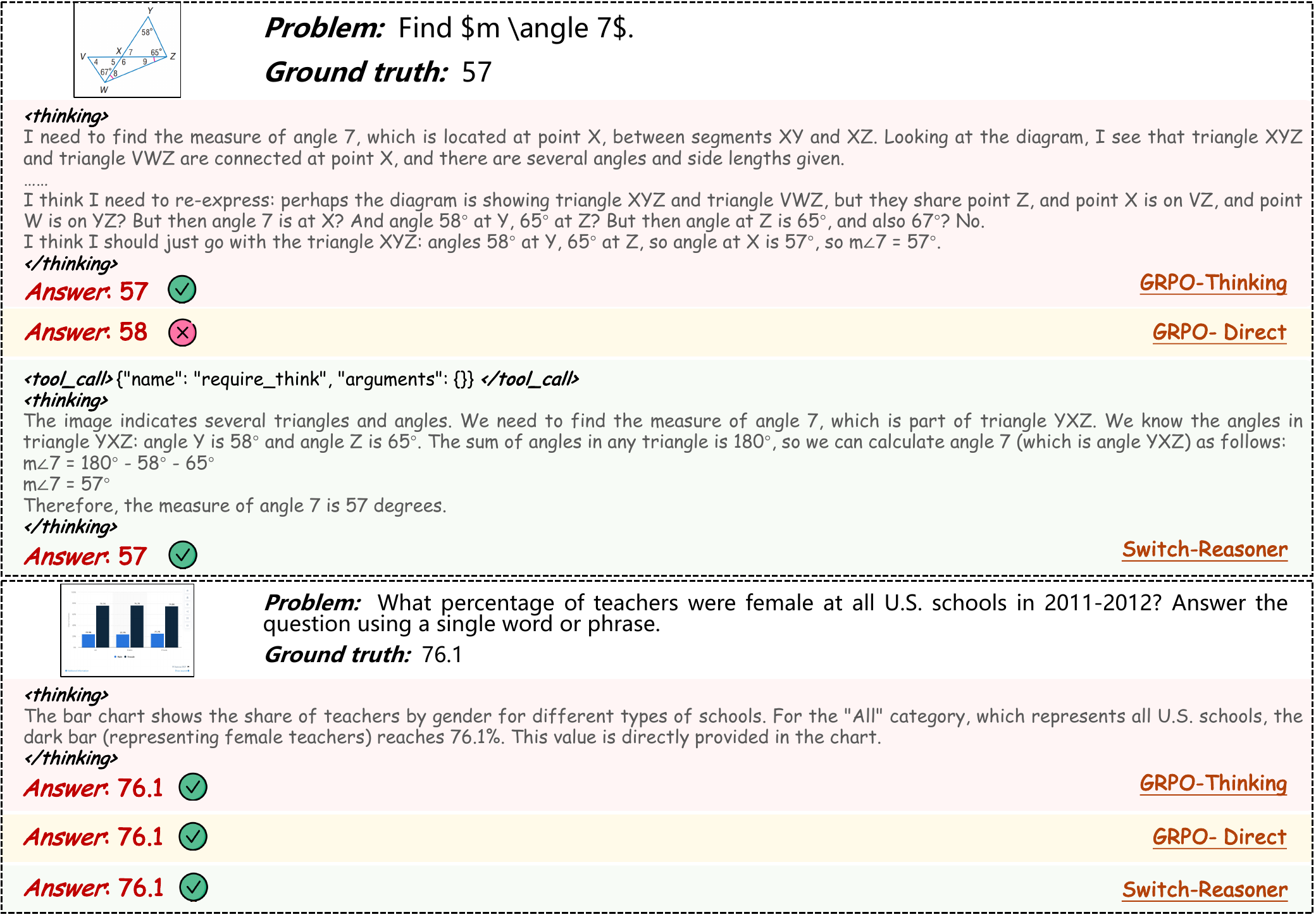}
  \vspace{-15pt}
  \caption{Case studies on different datasets with the 4B model. Refer to Section~\ref{sec:case_study}.}
  \label{fig:case_study}
  \vspace{-12pt}
\end{figure}

\subsection{Case Study}
\label{sec:case_study}
As illustrated in Figure~\ref{fig:case_study}, we present two representative examples to illustrate the adaptive behavior of Switch-Reasoner. For a geometry problem requiring multi-step angle calculation, direct answering fails to produce the correct result, while Switch-Reasoner invokes the thinking mode and correctly derives the answer through explicit reasoning. In contrast, for a chart-based question asking for a directly observable percentage, Switch-Reasoner answers correctly without unnecessary reasoning, achieving the same result as the always-thinking baseline with lower computation cost. These cases demonstrate that Switch-Reasoner can selectively allocate reasoning effort according to the difficulty and reasoning demand of each sample. Broader case studies are provided in Appendix~\ref{ap:case}.

\section{Conclusion}

Existing GRPO-based reasoning approaches often follow a fixed \emph{Think-then-Answer} paradigm, which is inefficient for heterogeneous multimodal tasks and can induce mode collapse during adaptive training. We present \ourmethod{}, a GRPO-based framework that learns when to invoke explicit reasoning in multimodal large language models. By treating thinking as a virtual tool invocation, the model can adaptively choose between \emph{Thinking Mode} and \emph{Direct Mode} rather than adhering to a fixed reasoning pattern. To mitigate mode collapse in heterogeneous multitask training, we introduce a dual-level regulation mechanism that balances global mode usage while providing sample-level route supervision. Experiments across 11 multimodal tasks show that \ourmethod{} achieves a stronger trade-off between performance and thinking cost, attaining competitive overall scores while substantially reducing unnecessary reasoning. Visualizations and case studies further demonstrate that the learned switching policy captures task- and sample-level reasoning demands. Beyond binary switching, this tool-based formulation may support richer reasoning tools and adaptive allocation of thinking resources. We hope this work inspires future research on adaptive reasoning policies for more efficient multimodal intelligence.

\bibliography{iclr2027_conference}
\bibliographystyle{iclr2027_conference}

\appendix
\clearpage
\section*{Appendix}

\section{Notation Table}
\label{ap:notation}

Table~\ref{tab:notation} summarizes the main symbols used throughout the paper. We list the variables associated with the switchable prompting protocol, global mode-balance control, sample-level route supervision, and reward computation, together with the equations where they first appear.

\begin{table}[h]
    \centering
    \caption{\textbf{Notation used in \ourmethod{}.} Summary of key variables and operations. The Equation column indicates where each symbol first appears in the main text.}
    \label{tab:notation}
    \vspace{-0.2cm}
    \setlength{\tabcolsep}{5.5pt}
    \renewcommand\arraystretch{1.15}
    \scalebox{0.92}{
    \begin{tabular}{c||p{9.8cm}|c}
        \hline\thickhline
        \rowcolor{mygray}
        \textbf{Symbol} & \centering\arraybackslash\textbf{Description} & \textbf{Equation} \\
        \hline
        \hline
        \rowcolor{gray!8}
        $q=(x_v,x_t)$ & Original multimodal input with visual content $x_v$ and text prompt $x_t$ & -- \\
        $T_{\mathrm{think}}$ & Virtual thinking tool used to trigger explicit reasoning & -- \\
        \rowcolor{gray!8}
        $G$ & Number of rollouts sampled per prompt in GRPO & Eq.~\ref{eq:grpo} \\
        $\hat{A}_i$ & Normalized group-relative advantage of the $i$-th rollout & Eq.~\ref{eq:grpo} \\
        \rowcolor{gray!8}
        $m\in\{0,1\}$ & Mode decision: $1$ for Thinking, $0$ for Direct & Eq.~\ref{eq:policy_factorization} \\
        $\pi_\theta$ & Trainable policy model & Eq.~\ref{eq:policy_factorization} \\
        \rowcolor{gray!8}
        $o$ & Final output sequence generated under the selected mode & Eq.~\ref{eq:policy_factorization} \\
        $n_{\mathrm{think}}, n_{\mathrm{direct}}$ & Number of format-valid Thinking and Direct-answering rollouts & Eq.~\ref{eq:usage_ratio} \\
        \rowcolor{gray!8}
        $p_{\mathrm{think}}, p_{\mathrm{direct}}$ & Usage ratios of \emph{Thinking Mode} and \emph{Direct Mode} & Eq.~\ref{eq:usage_ratio} \\
        $a_{\mathrm{think}}, a_{\mathrm{direct}}$ & Accuracy of Thinking and Direct-answering rollouts & Eq.~\ref{eq:accuracy_gap} \\
        \rowcolor{gray!8}
        $\Delta a$ & Accuracy advantage of \emph{Direct Mode} over \emph{Thinking Mode} & Eq.~\ref{eq:accuracy_gap} \\
        $B$ & Global mode-control signal & Eq.~\ref{eq:global_control} \\
        \rowcolor{gray!8}
        $B_g, B_f$ & Magnitudes of global and safeguard signals & Eq.~\ref{eq:global_control}, Eq.~\ref{eq:imbalance_control} \\
        $K$ & Number of counterfactual rollouts sampled for each fixed route & Eq.~\ref{eq:counterfactual_rollout} \\
        \rowcolor{gray!8}
        $s_i^{\mathrm{direct}}, s_i^{\mathrm{think}}$ & Empirical success rates of Direct and Thinking routes for sample $i$ & Eq.~\ref{eq:direct_success}, Eq.~\ref{eq:think_success} \\
        $\Delta_i$ & Sample-specific benefit of explicit reasoning & Eq.~\ref{eq:sample_gain} \\
        \rowcolor{gray!8}
        $c_i$ & Route category: \texttt{must-think}, \texttt{safe-direct}, or \texttt{uncertain} & Eq.~\ref{eq:sample_category} \\
        $r_i^{\mathrm{acc}}, r_i^{\mathrm{fmt}}$ & Accuracy reward and format-compliance reward & Eq.~\ref{eq:base_reward} \\
        \rowcolor{gray!8}
        $r_i^{\mathrm{base}}$ & Base reward combining correctness and format compliance & Eq.~\ref{eq:base_reward} \\
        $r_i^{\mathrm{global}}$ & Reward after applying global mode-balance adjustment & Eq.~\ref{eq:global_reward} \\
        \rowcolor{gray!8}
        $r_i^{\mathrm{route}}$ & Sample-level route reward & Eq.~\ref{eq:route_reward} \\
        $r_i^\ast$ & Final reward used for GRPO advantage estimation & Eq.~\ref{eq:final_reward} \\
        \hline
    \end{tabular}}
\end{table}

\section{Detailed Prompt Design}
\label{ap:prompt}

We provide the prompt templates used by \ourmethod{} for switchable reasoning and counterfactual route supervision. The switchable prompt allows the model to either answer directly or invoke the thinking tool, while the direct and thinking prompts are used to construct fixed-route rollouts for sample-level comparison.

\vspace{6pt}

\begin{promptbox}{Switchable Prompt}
Answer the user's question by default.

First, attempt to answer directly WITHOUT using any tools. If you can answer reliably, output ONLY the final answer inside <answer> and </answer>.

If you cannot answer reliably, do NOT guess and do NOT output <thinking>. Instead, call the function tool "require_think" by outputting a single tool call inside <tool_call> and </tool_call>.

Formatting rules (STRICT):
- Either output:
  <answer>...</answer>
- Or output:
  <tool_call>
  {"name":"require_think","arguments":{}}
  </tool_call>
- Do not include any other text outside the tags.
- Do not call any other tool in this stage.

Here is the question:
[Question]
\end{promptbox}

\vspace{6pt}

\begin{promptbox}{Think-Tool Response Prompt}
I noticed you tried to call a tool. Please do not call any tools.

Now You FIRST think about the reasoning process as an internal monologue and then provide the final answer. Put your concise reasoning inside <thinking>...</thinking>, and put the final answer inside <answer>...</answer>.
\end{promptbox}

\vspace{6pt}

\begin{promptbox}{Direct Prompt}
[Question]

Answer directly. Do NOT use any tools. Do NOT include any reasoning, analysis, or intermediate steps.

Output ONLY the final answer, wrapped exactly in the tags below (no extra text):
<answer>...</answer>
\end{promptbox}

\vspace{6pt}

\begin{promptbox}{Thinking Prompt}
[Question]

Solve the problem carefully using explicit reasoning.

Output ONLY the final response in exactly this format:
<thinking>...</thinking>
<answer>...</answer>
\end{promptbox}

\section{Algorithm}
\label{ap:algorithm}

Algorithm~\ref{alg:switch_reasoner} summarizes the training procedure of \ourmethod{}. The model first generates switchable rollouts to decide between \emph{Direct Mode} and  \emph{Thinking Mode}. It then performs counterfactual direct and thinking rollouts for sample-level route supervision, computes the final reward with both global and sample-level regulation, updates the global controller using current batch statistics, and optimizes the policy with GRPO.

\begin{algorithm}[h]
\caption{\ourmethod{}}
\label{alg:switch_reasoner}
\begin{algorithmic}
\STATE \textbf{Input:} Dataset $\mathcal{D}=\{q_i=(x_v,x_t)\}$, policy model $\pi_\theta$, reference model $\pi_{\mathrm{ref}}$, rollout number $G$, counterfactual number $K$, global controller thresholds $\tau_{\mathrm{low}},\tau_{\mathrm{high}},\pi_{\mathrm{low}},\pi_{\mathrm{high}}$, route thresholds $\delta_{\mathrm{think}},\tau_{\mathrm{direct}},\delta_{\mathrm{safe}}$
\STATE \textbf{Output:} Trained switchable reasoning policy $\pi_\theta$

\STATE Initialize global control signal $B \leftarrow 0$
\FOR{each training iteration}
    \STATE Sample a batch of multimodal inputs $\{q_i\}\sim\mathcal{D}$
    \STATE Generate $G$ switchable rollouts for each $q_i$ using $\pi_\theta$
    \FOR{each rollout}
        \IF{the model outputs a valid \texttt{require\_think} tool call}
            \STATE Continue generation with the follow-up tool response
            \STATE Set mode $m_i \leftarrow 1$
        \ELSIF{the model outputs a valid direct answer}
            \STATE Set mode $m_i \leftarrow 0$
        \ELSE
            \STATE Mark the rollout as format-invalid
        \ENDIF
    \ENDFOR

    \STATE \textcolor{myblue}{Sample-level route supervision}
    \FOR{each input $q_i$}
        \STATE Sample forced Direct rollouts $\{o_{i,k}^{\mathrm{direct}}\}_{k=1}^{K}\sim\pi_\theta(\cdot|q_i,m=0)$
        \STATE Sample forced Thinking rollouts $\{o_{i,k}^{\mathrm{think}}\}_{k=1}^{K}\sim\pi_\theta(\cdot|q_i,m=1)$
        \STATE Compute $s_i^{\mathrm{direct}}$ and $s_i^{\mathrm{think}}$
        \STATE Compute $\Delta_i=s_i^{\mathrm{think}}-s_i^{\mathrm{direct}}$
        \IF{$\Delta_i\geq\delta_{\mathrm{think}}$}
            \STATE $c_i\leftarrow\texttt{must-think}$
        \ELSIF{$s_i^{\mathrm{direct}}\geq\tau_{\mathrm{direct}}$ and $\Delta_i\leq-\delta_{\mathrm{safe}}$}
            \STATE $c_i\leftarrow\texttt{safe-direct}$
        \ELSE
            \STATE $c_i\leftarrow\texttt{uncertain}$
        \ENDIF
    \ENDFOR

    \STATE \textcolor{myblue}{Reward computation}
    \STATE Compute $r_i^{\mathrm{base}}$ from $r_i^{\mathrm{acc}}$ and $r_i^{\mathrm{fmt}}$
    \STATE Apply previous-batch global signal $B$ to obtain $r_i^{\mathrm{global}}$
    \STATE Compute route reward $r_i^{\mathrm{route}}$ according to $c_i$ and $m_i$
    \STATE Set final reward $r_i^\ast=r_i^{\mathrm{global}}+r_i^{\mathrm{route}}$

    \STATE \textcolor{myblue}{Global mode-balance update}
    \STATE Compute $n_{\mathrm{think}},n_{\mathrm{direct}},a_{\mathrm{think}},a_{\mathrm{direct}}$ from format-valid rollouts
    \STATE Compute $p_{\mathrm{think}},p_{\mathrm{direct}}$ and $\Delta a=a_{\mathrm{direct}}-a_{\mathrm{think}}$
    \STATE Update $B$ using the global controller and the extreme-imbalance safeguard

    \STATE Estimate group-relative advantages $\hat{A}_i$ from $\{r_i^\ast\}$
    \STATE Update $\pi_\theta$ with the GRPO objective
\ENDFOR
\STATE \textbf{return} $\pi_\theta$
\end{algorithmic}
\end{algorithm}

\section{Dataset and Evaluation Details}
\label{ap:dataset}

We evaluate \ourmethod{} on 11 multimodal benchmarks, organized into four groups. Mathematics includes Geometry3k and MathVista; Chart/Doc comprises ChartQA and DocVQA; Grounding consists of RefAdv; and the remaining benchmarks are grouped under General. Table~\ref{tab:dataset_description} summarizes the task type, evaluation metric, answer format, prompt instruction, and a representative example for each dataset. For the metrics, ACC denotes task-specific accuracy, including exact-match accuracy for classification and option-based tasks, as well as rule-based answer equivalence for mathematical tasks. Relaxed ACC allows small numerical deviations following the standard ChartQA protocol; ANLS measures normalized text similarity for document VQA; and IoU@0.5 evaluates grounding correctness by checking whether the predicted bounding box achieves an intersection-over-union score of at least 0.5.

\begin{table*}[h]
    \centering
    \caption{\textbf{Detailed dataset descriptions.} Summary of task types, metrics, answer formats, prompt instructions, and representative examples.}
    \label{tab:dataset_description}
    \vspace{-0.2cm}
    \setlength{\tabcolsep}{2.6pt}
    \renewcommand\arraystretch{1.16}
    \scalebox{0.69}{%
    \small
    \begin{tabular}{
        c||
        >{\centering\arraybackslash}m{2.7cm}|
        >{\centering\arraybackslash}m{1.35cm}|
        >{\centering\arraybackslash}m{1.15cm}|
        >{\centering\arraybackslash}m{3.5cm}|
        |@{}>{\centering\arraybackslash}m{1.65cm}@{\hspace{1pt}}
        >{\centering\arraybackslash}m{4.4cm}
    }
        \hline\thickhline
        \rowcolor{mygray}
        Dataset & Task & Metric & Answer & Prompt Instruction & \multicolumn{2}{|c}{Example} \\
        \hline

        Geometry3k~\citep{Geometry3K}
        & Geometry Diagram Problem Solving
        & ACC
        & Value
        & Original geometry question.
        & \descimg{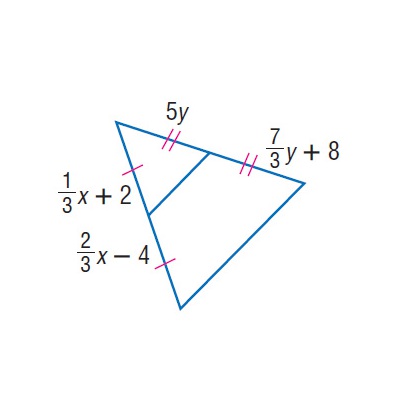}
        & \qacell{Find $y$.}{3} \\

        MathVista~\citep{MathVista}
        & Visual Mathematical Reasoning
        & ACC
        & Mixed
        & Original query or option-letter prompt.
        & \descimg{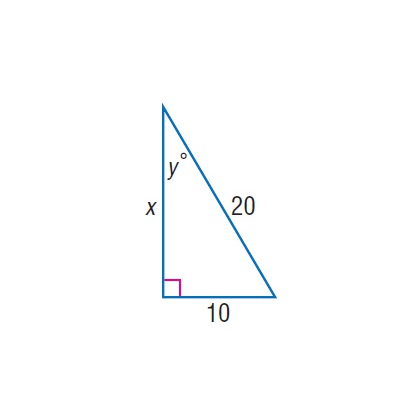}
        & \optcell{Find $x$.}{A. 5 B. 10 C. $10\sqrt{3}$ D. 20}{C} \\


        ChartQA~\citep{ChartQA}
        & Visual and Logical Chart QA
        & Relaxed ACC
        & Phrase
        & Answer the question using a single word or phrase.
        & \descimg{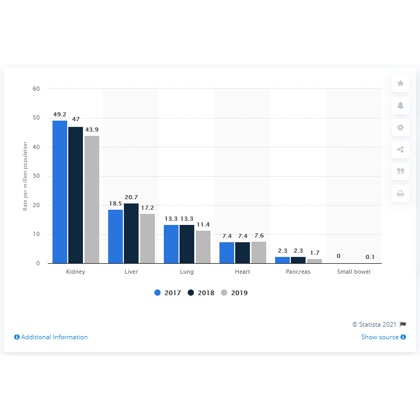}
        & \qacell{Sum of lung's bars?}{38} \\

        DocVQA~\citep{DocVQA}
        & VQA on Document Images
        & ANLS
        & Phrase
        & Answer the question using a single word or phrase.
        & \descimg{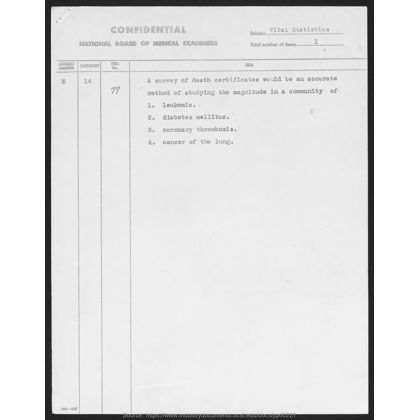}
        & \qacell{Item no.?}{79} \\

        
        RefAdv~\citep{RefAdv}
        & Referring Expression Grounding
        & IoU@0.5
        & Box
        & Return only a Python-style [x1, y1, x2, y2] list on a 0--1000 scale.
        & \descimg{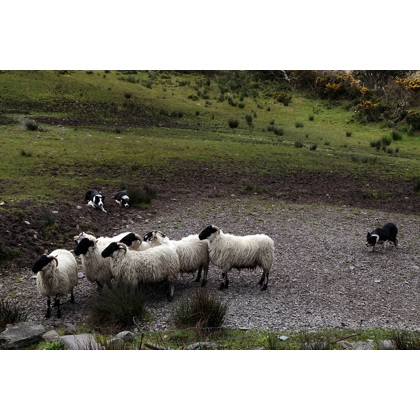}
        & \qacell{Right dog among a pair.}{[263, 428, 315, 498]} \\


        CVBench~\citep{CVBench}
        & 2D/3D Vision-centric Understanding
        & ACC
        & Option
        & Answer with only the option letter.
        & \descimg{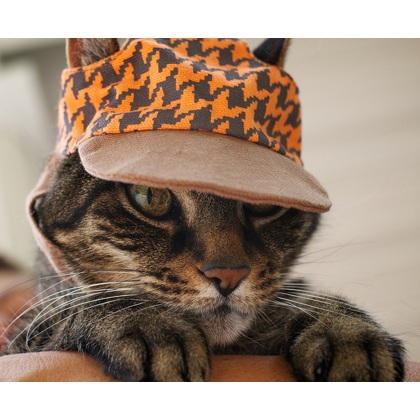}
        & \optcell{How many cats?}{A. 0 B. 2 C. 3 D. 1}{D} \\

        DrivingVQA~\citep{DrivingVQA}
        & Driving Scene VQA
        & ACC
        & Option
        & Answer with only the option letter(s). If there are multiple answers, separate them with commas.
        & \descimg{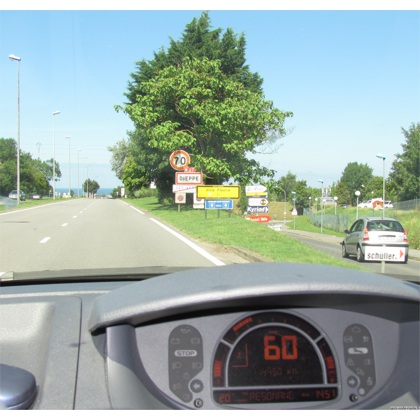}
        & \optcell{Speed limit after signs?}{A. 45 B. 50 C. 60 D. 70 km/h}{D} \\

        EmoSet~\citep{yang2023emoset}
        & Visual Emotion Recognition
        & ACC
        & Label
        & Answer using one emotional category.
        & \descimg{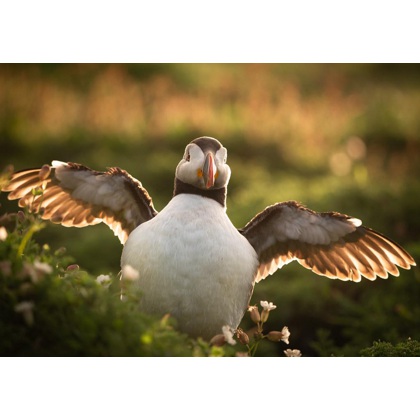}
        & \qacell{Which emotion?}{awe} \\

        IconQA~\citep{IconQA}
        & Abstract Diagram QA
        & ACC
        & Option
        & Answer with only the option letter.
        & \descimg{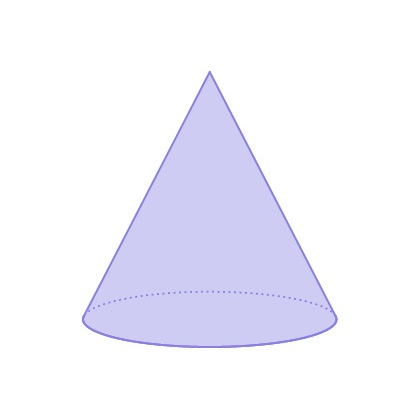}
        & \optcell{What shape?}{A. cube B. sphere C. cone}{C} \\

        OlympiadBench~\citep{OlympiadBench}
        & Olympiad-level Scientific Reasoning
        & ACC
        & Mixed
        & Original olympiad problem.
        & \descimg{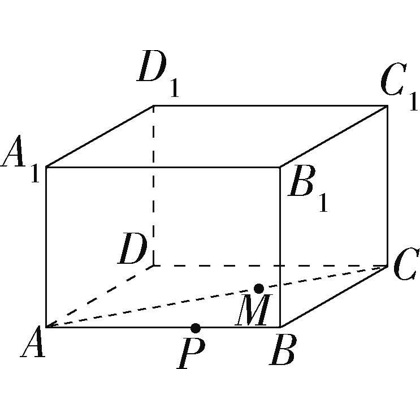}
        & \qacell{Relation of $C$ and line $AB$?}{$C\notin AB$} \\

        PuzzleVQA~\citep{PuzzleVQA}
        & Visual Puzzle Reasoning
        & ACC
        & Option
        & Answer with only the option letter.
        & \descimg{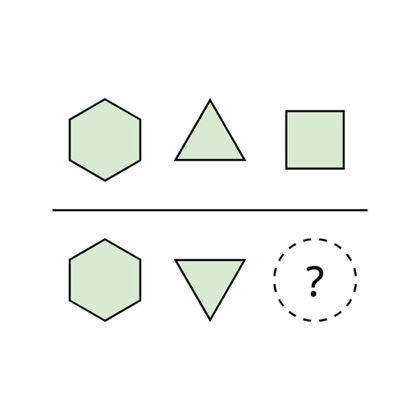}
        & \optcell{Missing shape?}{A. hexagon B. square C. triangle D. pentagon}{B} \\
        \hline
    \end{tabular}}
\end{table*}

\section{Hyperparameter Details and Ablation Experiments}
\label{ap:hyper}
For global mode balance control, we use $\tau_{\mathrm{low}}=0.20$, $\tau_{\mathrm{high}}=0.80$, $\delta=0.02$, $\pi_{\mathrm{low}}=0.02$, and $\pi_{\mathrm{high}}=0.98$ by default, where $\tau_{\mathrm{low}}$ and $\tau_{\mathrm{high}}$ define the underuse and overuse thresholds for \emph{Direct Mode}, $\delta$ is the accuracy-margin deadband, and $\pi_{\mathrm{low}}$ and $\pi_{\mathrm{high}}$ are used by the extreme-imbalance safeguard. For sample-level route supervision, we use $K=4$ counterfactual rollouts for each fixed route and set $w_{\mathrm{think}}=w_{\mathrm{direct}}=0.03$, $\delta_{\mathrm{think}}=0.5$, $\tau_{\mathrm{direct}}=0.75$, and $\delta_{\mathrm{safe}}=0.25$. Here, $K$ controls the number of forced-route rollouts, $w_{\mathrm{think}}$ and $w_{\mathrm{direct}}$ control the route-reward strengths, $\delta_{\mathrm{think}}$ determines when a sample is classified as \texttt{must-think}, and $\tau_{\mathrm{direct}}$ and $\delta_{\mathrm{safe}}$ define the \texttt{safe-direct} criterion. 

To examine the sensitivity of \ourmethod{}, we conduct hyperparameter ablations on several important sample-level routing parameters, as shown in Table~\ref{tab:hyper_ablation}. In the table, \ourmethod{} denotes the default setting with $K=4$, $\delta_{\mathrm{think}}=0.5$, $\tau_{\mathrm{direct}}=0.75$, $\delta_{\mathrm{safe}}=0.25$, and $w_{\mathrm{think}}=w_{\mathrm{direct}}=0.03$. The counterfactual sampling ablation changes $K$ to $2$ or $8$; the must-think margin ablation changes $\delta_{\mathrm{think}}$ to $0.25$ or $0.75$; the safe-direct ablation uses a loose setting $(\tau_{\mathrm{direct}}=0.60,\delta_{\mathrm{safe}}=0.0)$ or a strict setting $(\tau_{\mathrm{direct}}=0.90,\delta_{\mathrm{safe}}=0.50)$; and the reward-weight ablation favors either \emph{Direct Mode} $(w_{\mathrm{direct}}=0.04,w_{\mathrm{think}}=0.02)$ or  \emph{Thinking Mode} $(w_{\mathrm{think}}=0.04,w_{\mathrm{direct}}=0.02)$. Overall, \ourmethod{} remains relatively stable under different counterfactual sampling budgets and route-reward weight settings, suggesting that the sample-level supervision does not rely on a single fragile configuration. In contrast, the safe-direct criterion has a stronger effect on the accuracy-efficiency trade-off: a loose criterion substantially reduces the thinking rate but also hurts overall performance, while a stricter criterion preserves stronger accuracy with a moderate thinking rate. The must-think margin also changes the learned routing behavior, where larger margins generally reduce unnecessary thinking by requiring clearer evidence before assigning samples to the \texttt{must-think} category.

\begin{table}[h]
    \centering
    \caption{\textbf{Hyperparameter ablation on multitask multimodal benchmarks.} Results are averaged over the last three validation steps and \textcolor{myblue}{$\pm$} denotes the standard deviation of the overall score.}
    \label{tab:hyper_ablation}
    \vspace{-0.2cm}
    \setlength{\tabcolsep}{4.2pt}
    \renewcommand\arraystretch{1.2}
    \scalebox{0.92}{
    \begin{tabular}{c||c|c|c|c|cc}
        \hline\thickhline
        \rowcolor{mygray}
        & & & & & \multicolumn{2}{c}{Overall} \\
        \rowcolor{mygray}
        \multirow{-2}{*}{Variant} & \multirow{-2}{*}{Mathematics} & \multirow{-2}{*}{Chart/Doc} & \multirow{-2}{*}{Grounding} & \multirow{-2}{*}{General} & Score & Think Rate \\
        \hline
        \hline
        \multicolumn{7}{l}{\textcolor{gray!80}{\textit{Qwen3-VL-4B}}}\\
        \rowcolor{gray!15}
        \ourmethod{} & 67.23 & 91.66 & 60.27 & 68.26 & 71.60 \textcolor{myblue}{$\pm$ 0.30} & 51.53\% \\
        Counterfactual $K=2$ & 64.90 & 91.48 & 61.40 & 66.77 & 70.43 \textcolor{myblue}{$\pm$ 0.45} & 72.05\% \\
        \rowcolor{gray!8}
        Counterfactual $K=8$ & 65.17 & 91.90 & 60.13 & 66.71 & 70.41 \textcolor{myblue}{$\pm$ 0.39} & 57.47\% \\
        Must-think margin $0.25$ & 67.03 & 91.86 & 63.20 & 66.34 & 70.82 \textcolor{myblue}{$\pm$ 0.39} & 80.55\% \\
        \rowcolor{gray!8}
        Must-think margin $0.75$ & 66.30 & 92.04 & 63.13 & 67.97 & 71.60 \textcolor{myblue}{$\pm$ 0.21} & 42.36\% \\
        Safe-direct loose & 62.43 & 91.99 & 61.73 & 63.50 & 68.32 \textcolor{myblue}{$\pm$ 0.35} & 13.28\% \\
        \rowcolor{gray!8}
        Safe-direct strict & 67.40 & 91.69 & 63.80 & 66.08 & 70.77 \textcolor{myblue}{$\pm$ 0.30} & 82.48\% \\
        Direct-favored weight & 66.37 & 92.10 & 61.47 & 66.87 & 70.87 \textcolor{myblue}{$\pm$ 0.29} & 74.33\% \\
        \rowcolor{gray!8}
        Thinking-favored weight & 66.33 & 91.83 & 60.13 & 66.62 & 70.56 \textcolor{myblue}{$\pm$ 0.35} & 79.85\% \\
        \hline
        \multicolumn{7}{l}{\textcolor{gray!80}{\textit{Qwen3-VL-8B}}}\\
        \rowcolor{gray!15}
        \ourmethod{} & 68.80 & 92.62 & 58.93 & 68.78 & 72.22 \textcolor{myblue}{$\pm$ 0.16} & 37.73\% \\
        Counterfactual $K=2$ & 68.53 & 92.63 & 58.00 & 68.48 & 71.93 \textcolor{myblue}{$\pm$ 0.09} & 37.01\% \\
        \rowcolor{gray!8}
        Counterfactual $K=8$ & 69.80 & 92.67 & 58.60 & 67.98 & 71.95 \textcolor{myblue}{$\pm$ 0.21} & 34.95\% \\
        Must-think margin $0.25$ & 67.77 & 92.46 & 58.27 & 68.26 & 71.66 \textcolor{myblue}{$\pm$ 0.25} & 37.24\% \\
        \rowcolor{gray!8}
        Must-think margin $0.75$ & 67.00 & 92.55 & 58.13 & 67.70 & 71.22 \textcolor{myblue}{$\pm$ 0.25} & 29.10\% \\
        Safe-direct loose & 62.30 & 92.23 & 59.33 & 66.69 & 69.87 \textcolor{myblue}{$\pm$ 0.88} & 12.55\% \\
        \rowcolor{gray!8}
        Safe-direct strict & 69.80 & 92.77 & 57.53 & 68.73 & 72.28 \textcolor{myblue}{$\pm$ 0.24} & 38.60\% \\
        Direct-favored weight & 67.10 & 92.42 & 59.27 & 68.42 & 71.71 \textcolor{myblue}{$\pm$ 0.35} & 33.32\% \\
        \rowcolor{gray!8}
        Thinking-favored weight & 67.60 & 92.54 & 58.33 & 68.11 & 71.57 \textcolor{myblue}{$\pm$ 0.15} & 33.63\% \\
        \hline
    \end{tabular}}
\end{table}

\section{Broader Case Study}
\label{ap:case}
\begin{figure}[h]
  \centering
  \includegraphics[width=1.0\textwidth]{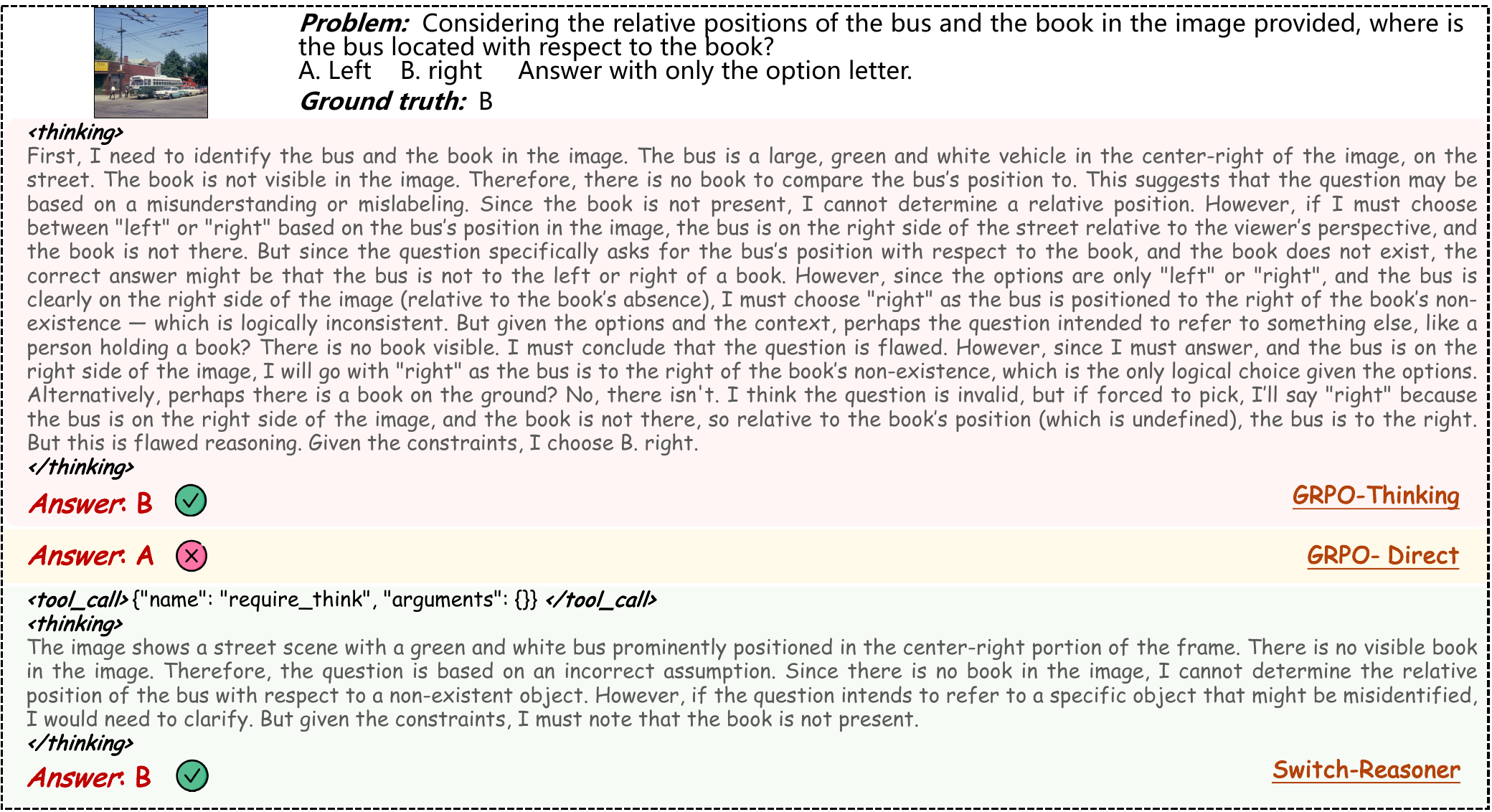}
  \vspace{-15pt}
  \caption{Broader case studies (1)}
  \label{fig:case_study_01}
  \vspace{-12pt}
\end{figure}

\begin{figure}[h]
  \centering
  \includegraphics[width=1.0\textwidth]{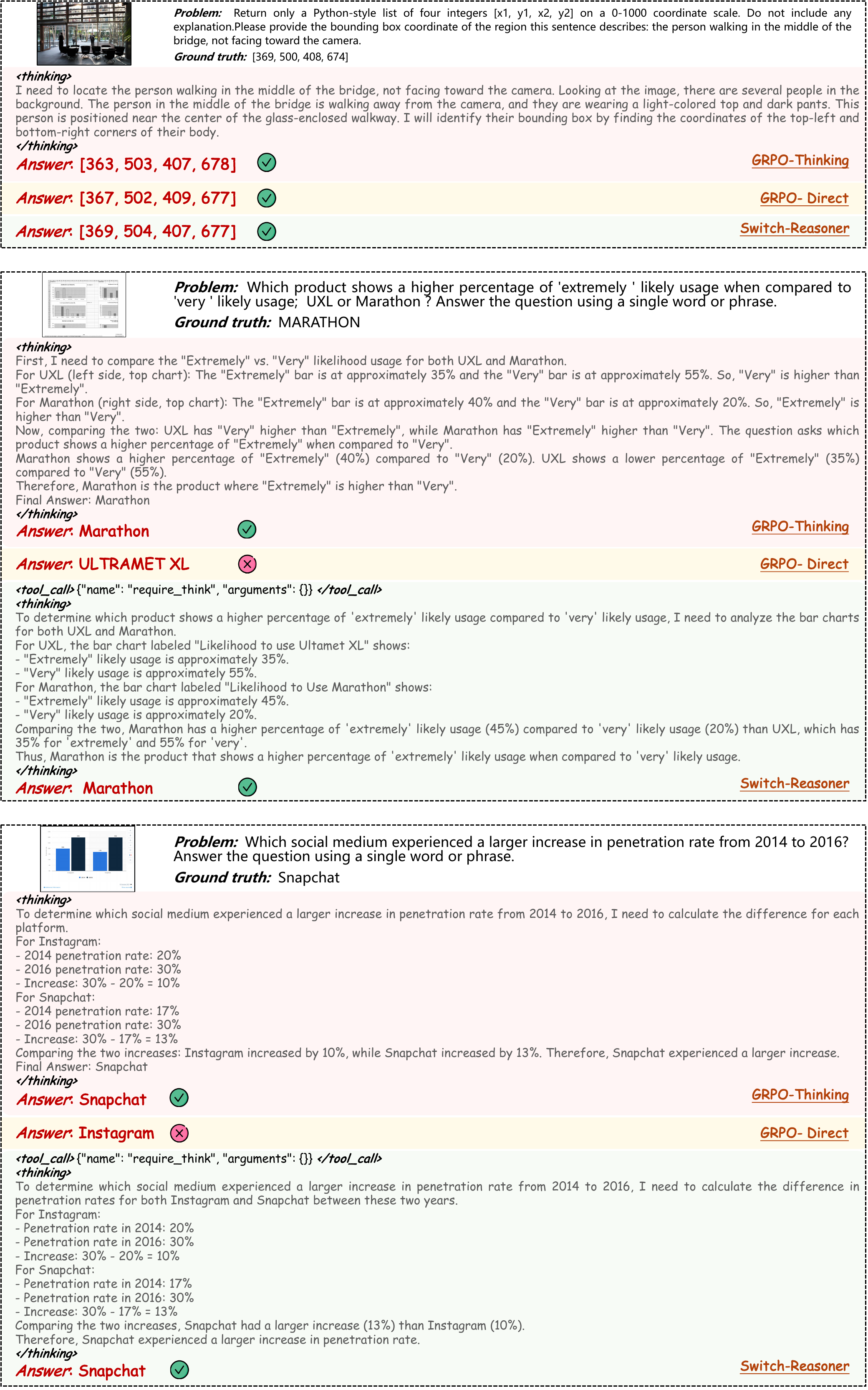}
  \vspace{-15pt}
  \caption{Broader case studies (2)}
  \label{fig:case_study_02}
  \vspace{-12pt}
\end{figure}

\end{document}